%% file: main.tex
\documentclass[runningheads]{llncs}
\usepackage[T1]{fontenc}
\usepackage{graphicx}
\usepackage{url}
\usepackage{algorithm}
\usepackage{arevmath}     
\usepackage[noend]{algpseudocode}
\usepackage{caption}
\usepackage{subcaption}
\usepackage{amsmath}
\usepackage[dvipsnames]{xcolor}
\definecolor{wikiblue}{HTML}{0645AD}
\usepackage{multicol}
\newcommand{\citationneeded}[1][]{\textsuperscript{\color{blue} [citation needed]}}

\begin{document}
\title{Named Entity Recognition for Partially Annotated Datasets}
\titlerunning{Named Entity Recognition for Partially Annotated Datasets}
%
 \author{Michael~Strobl\inst{1} \and Amine Trabelsi\inst{2} \and Osmar~Za\"iane\inst{1}}
 \authorrunning{M. Strobl et al.}
 \institute{University of Alberta, Edmonton, Canada\\
 \email{$\{$mstrobl,zaiane$\}$@ualberta.ca}
 	\and
 Lakehead University, Thunder Bay, Canada\\
 \email{atrabels@lakeheadu.ca}}
\maketitle              
\begin{abstract}
The most common Named Entity Recognizers are usually sequence taggers trained on fully annotated corpora, i.e. the class of all words for all entities is known. Sequence taggers are fast to train and to make predictions. Partially annotated corpora, i.e. some but not all entities of some types are annotated, are too noisy for training sequence taggers since the same entity may be annotated one time with its true type but not another time, misleading the tagger. Therefore, we are comparing three training strategies for partially annotated datasets and an approach to derive new datasets for new classes of entities from Wikipedia without time-consuming manual data annotation. In order to properly verify that our data acquisition and training approaches are plausible, we manually annotated test datasets for two new classes, namely food and drugs, and report the resulting performance of all trained models on these test datasets.

\end{abstract}
%

\section{Introduction}
\label{sec:introduction}
\input{sections/1_introduction}

\section{Related Work}
\label{sec:related}
\input{sections/2_related_works}

\section{Method}
\label{sec:method}
\input{sections/3_method}

\section{Evaluation}
\label{sec:evaluation}
\input{sections/4_evaluation}


\section{Conclusion}
\label{sec:conclusion}
\input{sections/6_conclusion}

\end{document}

%% file: sections/1_introduction.tex

Named Entity Recognition (NER) is one of the most popular tasks in NLP. The goal for NER is to classify each token (usually corresponding to a word) in a sequence of tokens according to a scheme and a set of classes. The set of classes a model can recognize is dependent on the dataset it is trained on, e.g. \textit{Person}, \textit{Organization}, \textit{Location} and \textit{Miscellaneous} for the popular CoNLL 2003 NER dataset \cite{sang2003introduction}. Typically a sequence tagging model (e.g. \cite{lample-etal-2016-neural} or \cite{devlin2019bert}) is used to achieve this task, which takes a sentence as a sequence of tokens as input and outputs a sequence of classes all at once.

In order to train such a model, high-quality manually annotated data is necessary with each entity in the dataset assigned its correct class. Datasets with partial annotations, mainly without all entities being annotated, are noisy and lead to worse model performance. This is due to the fact that entities in the dataset, which are not annotated as such, are automatically considered as belonging to the \textit{Outside} class, i.e. tokens not belonging to an entity. Therefore, the model is trained not to recognize these, even though it should, leading to a model which may be tempted to ignore a certain number of entities when used for making predictions.

However, partially annotated data is often easier to come by, e.g. through using hyperlinks from Wikipedia\footnote{\url{https://www.wikipedia.org/}} as entities. This is especially useful if classes other than the ones seen in common NER datasets are of interest. Furthermore, intuitively, why should it be necessary to annotate every single entity in a dataset to make a model learn this task? Humans are perfectly able to recognize mentions of an entity consistently after having it ``classified'' once. Consider the following sentence with partial links for animals from Wikipedia\footnote{\url{https://en.wikipedia.org/wiki/Animal_cracker}}:

\begin{quote}
    "Stauffer's animal crackers include bear, bison, camel, \textbf{\textcolor{wikiblue}{cow}}, \textbf{\textcolor{wikiblue}{cat}}, \textbf{\textcolor{wikiblue}{donkey}}, elephant, hippopotamus, horse, lion, \textbf{\textcolor{wikiblue}{mountain goat}}, rhinoceros, and tiger."
\end{quote}

This sentence originally contains four links to animals, while the other nine animals are not linked. Wikipedia often contains partially annotated sentences due to a Wikipedia policy discouraging editors either from linking the same article more than once or linking popular articles all together since they would not provide any new information to the reader. However, if the goal is to train a model, which is capable of recognizing the class ``Animal'', being able to use such data without manual annotations would simplify the task significantly.




Therefore, this paper aims to describe how it is possible to train commonly used models for NER on partially annotated datasets for new classes, without a significant manual effort for data annotation.

These are our main contributions:

\begin{itemize}
    \item Describing a procedure on how we can create partially annotated datasets for new classes derived from Wikipedia categories semi-automatically.
    \item Providing and comparing training strategies for NER models on partially annotated datasets.
    \item Releasing two manually annotated dataset of 500 sentences each for the classes \textit{Food} and \textit{Drugs} in order to test how generalizable our data extraction techniques are.
\end{itemize}

The remainder of this paper is outlined as follows: Section \ref{sec:related} shows some related work on how the problem of training models on partially annotated datasets has been approached before. We propose our method for data extraction from Wikipedia and model training strategies in Section \ref{sec:method}. The experimental evaluation can be found in Section \ref{sec:evaluation} with 
a conclusion in Section \ref{sec:conclusion}.

%% file: sections/2_related_works.tex

Jie et al. \cite{jie2019better} proposed an approach to train a BiLSTM-CRF model on partially annotated datasets. Their iterative approach tried to find the most likely labelling sequence that is compatible with the existing partial annotation sequence, i.e. the model is supposed to learn to assign the highest weight to the most likely (ideally correct) labelling sequence.

The CoNLL 2003 dataset for English was used for the evaluation (in addition to the CoNLL 2002 dataset for Spanish NER \cite{tjong-kim-sang-2002-introduction}). 50\% of the labelled entities were removed for testing their model, effectively lowering the recall when trained on this dataset. The best model achieved a 1.4\% F1-score reduction on CoNLL 2003 (compared to the same model architecture trained on the complete dataset without any entities removed). Only fully annotated (yet artificially perturbed) datasets were considered for the evaluation. While it would be technically possible to use partially annotated datasets with non-entity annotations\footnote{These are annotations indicating that this is knowingly not an entity, which is possible for partially annotated datasets derived from Wikipedia, which is described in more detail in the next section.} through ruling out some potential labelling sequences, 
this
was not tested. 


A different approach was proposed by Mayhew et al. \cite{mayhew2019named} for training BiLSTM-CRF models on existing datasets for a variety of languages, e.g. the popular CoNLL 2003 dataset for English NER \cite{sang2003introduction}. They used an iterative approach in order to learn a weight between 0.0 and 1.0 for each token, depending on whether the corresponding prediction should add to the loss. Whenever a span of tokens representing an entity is considered as non-entity, the weight should be close to 0.0, and in case of a proper entity annotation, the weight should be 1.0. The dataset was artificially perturbed to reduce precision as well as recall, i.e. to lower recall some entity annotations were removed and to lower precision some random spans of tokens were annotated as entities. Therefore, instead of trying to label the training sequence correctly, they tried to figure out which tokens are of class \textit{Outside} with high confidence (weight = 1.0) and which ones are probably entities (weight = 0.0) that should not add to the loss.

Their best model still suffered from an F1-score reduction of 5.7\% and in the same experiments the models from \cite{jie2019better} had an 8\% reduction. Note that the dataset also contained random spans of tokens added as entities, which was not tested by \cite{jie2019better}. Although it is not known which mistakes can be attributed to lowering recall or precision in the training dataset. The obvious drawback of their approach is that false negatives (entities in the dataset without annotations) are not supposed to be considered for training, effectively reducing the set of entities the model can be trained on. Only the weights for each token are adjusted, but not the labels. Furthermore, assuming a partially annotated dataset contains entities of some class as well as non-entity annotations of this class, i.e. entities that are known not to belong to this class (but it may not be known which class they actually belong to), their model cannot take advantage of this kind of information, it simply tries not to use these annotations for training. 

In addition, both aforementioned model architectures seem to be outdated for today's standards as models based on the Transformer architecture \cite{NIPS2017_3f5ee243}, specifically the pre-trained BERT model \cite{devlin2019bert}, achieve a significantly higher F1-score than BiLSTM-CRF models when trained on unperturbed datasets, e.g. see the results on CoNLL 2003 from \cite{devlin2019bert} compared to the popular LSTM-based approach from \cite{lample-etal-2016-neural}, which was specifically developed for NER. Both models are not tested on new datasets with new entity classes.

%% file: sections/3_method.tex

This section proposes our approach to create partially annotated NER datasets from Wikipedia for new classes and strategies to train models on these datasets without sacrificing prediction performance on entities from existing classes.

\subsection{Data creation}

When creating datasets for new classes for NER there are two problems to solve:

\begin{enumerate}
    \item Where can we get text data from? Entities of the class of interest maybe less abundant than common classes, such as \textit{Person}, \textit{Location} or \textit{Organization}. If simply random sentences, e.g. from the web, are included, the fraction of useful token spans maybe quite low\footnote{Mainly those token spans are useful that could potentially be part of the current class of interest, but are not always, e.g. ``Tomato'' could refer to the class Food or it could be a musician: \url{https://en.wikipedia.org/wiki/Tomato_(musician)}}.
    \item How can we annotate relevant token spans? Manual data annotation is a time-consuming task, even if done partially, which should be avoided. In addition, if a token span looks like a relevant entity, does it make sense to annotate it as well as non-entity for the class of interest? This mainly depends on whether a model can take advantage of that, i.e. whether the model can be told not to predict this class. The approach proposed by \cite{mayhew2019named} would not be able to.
\end{enumerate}

Wikipedia as a whole can be considered as a partially annotated dataset. Hyperlinks in articles correspond to entities and the hierarchical category system can be considered as a class hierarchy, which can be used to classify entities. Each article can have several categories attached by editors, although not all of them refer to a class in an NER-sense. For example, the article ``Salt''\footnote{\url{https://en.wikipedia.org/wiki/Salt}} has the following categories attached: ``Edible Salt'', ``Food additives'', ``Sodium minerals'' and ``Objects believed to protect from evil''. In case the class of interest is ``Food'', we can consider articles in the first two categories as relevant, whereas the latter two categories do not necessarily refer to food-related articles. Therefore, we only need to know which categories are relevant for this class in order to extract a set of articles and sentences they are linked to create a partially annotated dataset.

Algorithm \ref{alg:bfs} outlines a simple procedure to extract categories and articles from the Wikipedia category hierarchy using Breadth-first-search. We start at a base-category, e.g. ``Category:Food and Drink''\footnote{\url{https://en.wikipedia.org/wiki/Category:Food_and_drink}} for the class \textit{Food}. At each iteration of the loop in line 3 a category and 10 articles in this category (if available) are presented to the user and they have to decide whether to keep the category (and potentially the articles as well\footnote{Sometimes intermediate categories the user wants to keep contain uninteresting articles, in which case these articles can be ignored.}). If it is kept, all sub-categories (if available) are added to the queue. Ultimately, all categories the user wants to keep including all articles in these categories (if not explicitly excluded) are considered for the class of interest, and text from Wikipedia can be extracted.

\begin{algorithm}
\caption{Extract articles and sub-categories}
\label{alg:bfs}
\hspace*{\algorithmicindent} \textbf{Input} Wikipedia type hierarchy. \\
 \hspace*{\algorithmicindent} \textbf{Output} Partial hierarchy corresponding to entity class of interest.
 \begin{algorithmic}[1]

\Procedure{Bfs}{$cat_{start}$}
    \State queue = [$cat_{start}$]
    \While{queue not empty}
        \State $cat_{current} = queue.pop(0)$
        \State print $cat_{current}$ and 10 articles
        \State $input_{user}$ = input()
        \If{$input_{user} = 'y'$ or $input_{user}n = 's'$}
            \For {$cat_{sub}$ in $cat_{current}$}
                \State queue.append($cat_{sub}$)
                \If{$input_{user} = 'y'$}
                    \State Keep all articles in $cat_{current}$
                \EndIf
            \EndFor
        \EndIf
    \EndWhile
\EndProcedure

\end{algorithmic}
\end{algorithm}

These categories were added by the editors of Wikipedia and are often redundant, therefore it may be necessary to restart the procedure to avoid adding too many categories. However, in our experience it seems to be possible to finish it within 1 to 2 hours, at least for our test classes.

Since the training corpus for the new class $C$ should be somewhat difficult for a model to be trained on, it is also necessary to consider articles, which share aliases with articles in $C$ (many entity mentions can refer to entities of different types). This can be done with an alias dictionary derived from Wikipedia hyperlinks. We used the parser from \cite{strobl2020wexea} including their alias dictionary in order to extract sentences from Wikipedia, which contain hyperlinks of articles in $C$ (annotated as entities of class $C$) or hyperlinks of other articles that share aliases with those in $C$ (annotated as non-entities). This will result in a partially annotated corpus of entities and non-entities of class $C$, i.e. a set of sentences can be extracted from Wikipedia that contain relevant entities identified through their hyperlinks. 

In addition, the alias dictionary is used to annotate potential entities with an unknown type since, as we pointed out, not all entities are annotated in Wikipedia. Depending on the model, these entities would be excluded from training since the type is unknown. This is applied to all datasets used for training, e.g. CoNLL 2003 may contain entities of type \textit{Food}, which can be found and potentially excluded this way. Since an NER model should be trained on this kind of dataset as well as other datasets, such as CoNLL 2003, the CoreNLP NER \cite{manning-EtAl:2014:P14-5} is used to find all other entities of type \textit{Person}, \textit{Location}, \textit{Organization} and \textit{Miscellaneous}. These additional entities are also considered as non-entities, although since they are not gold-standard entities, we do not use their types otherwise for training, i.e. if a Location was found this way, the model is only trained not to recognize it as one of the new classes (if applicable), but it is not trained to recognize it as Location.


\subsection{Model training strategies}

When partially annotated data is introduced, the main goals for model training are:

\begin{enumerate}
    \item Classification accuracy for entities of existing classes should not suffer from the introduction of new data: We still rely on the CoNLL 2003 dataset for NER with classes \textit{Person}, \textit{Location}, \textit{Organization} and \textit{Miscellaneous}. Entities of these classes will very likely appear in the new datasets, but it is unknown where they are mentioned. A sequence tagger used for NER considers all tokens outside of the spans of entities as token of class \textit{Outside}. Therefore, these unlabeled mentions could mislead the classifier if trained on, leading to a decreased classification accuracy for existing classes.
    \item Similarly, a new entity class is introduced through partially annotated data and predictions by a model trained on such data should still be of high quality: A common multi-label single-class classifier, as used for NER in \cite{devlin2019bert} or \cite{lample-etal-2016-neural}, needs a label for each token. But in some cases such a label can either not be provided at all or it contains only partial information about the class, i.e. that it is not part of a specific class, but the class membership is unknown otherwise (a non-entity for a certain class). If in doubt, the \textit{Outside} class could be used here, leading to the aforementioned problem, or the tokens within the span of these entities should be excluded from training. But since this should be valuable information for the classifier, a model, that can take advantage of these mentions, is desirable.
\end{enumerate}

In the following, we propose multiple strategies for NER with partially annotated data with an evaluation in Section \ref{sec:evaluation} showing how well they meet these goals.


\begin{figure}
\centering
  \includegraphics[width=0.7\textwidth]{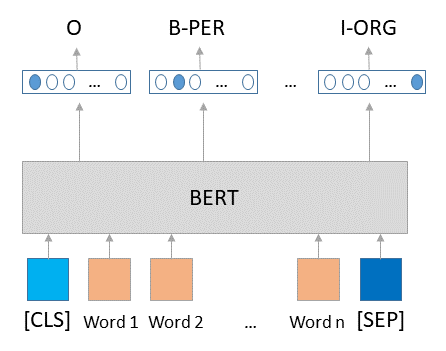}
  \caption{NER with BERT. Each token is embedded with the BERT model, which is used as input for the final classification layer.}
  \label{fig:bert_ner}
\end{figure}

\subsubsection{Ordinary sequence tagger}

The model described here is used as starting point for all subsequent models. We are using the NER model proposed in \cite{devlin2019bert}, which was published with the ubiquitous BERT model and shown in Figure \ref{fig:bert_ner}. An output layer with a softmax activation is used for token classification. The Categorical Cross-Entropy is used as loss function:

\begin{equation}
     J_1(\Theta) = -\frac{1}{N} \sum_{n=1}^N\sum_{i=1}^M y_i^{(n)} \log \hat{y}_i^{(n)}
\end{equation}

It is straightforward that this loss function is not capable of taking advantage of non-entities as well as entities of unknown type in the training dataset. Therefore, these entities are considered to belong to the \textit{Outside} class for this model, which is denoted as \textit{Softmax}-model.

\subsubsection{Sequence tagger ignoring entities}

If a span of tokens was annotated as non-entity or an entity of unknown type, it is probably detrimental for the model to simply classify it as \textit{Outside}. Therefore, through adjusting the loss function to ignore these tokens when training, we do not harm the model. This can be done with the following loss:

\begin{equation}
    J_2(\Theta) = -\frac{1}{N} \sum_{n=1}^N w^{(n)} \sum_{i=1}^M y_i^{(n)} \log \hat{y}_i^{(n)}
\end{equation}

A weight $w$ is added for each token with $w$ = 0 for non-entity tokens and tokens of unknown type, the model should not be trained on, and $w$ = 1 for all other tokens. This model is denoted as \textit{Softmax (weighted)}.


\subsubsection{Sequence tagger trained on non-entities}

So far the problem of taking advantage of the fact that some spans of tokens are known to be entities not belonging to the new class $C$, but unknown which is the true class, has not been solved yet. In order to do so, our model can be slightly adapted through using a multi-class multi-label approach with a sigmoid activation for each class in the output layer. The following loss function based on binary cross-entropy can be used here:

\begin{equation}
        J_3(\Theta) = -\frac{1}{N} \sum_{n=1}^N \sum_{i=1}^M w_i^{(n)}
        (y_i^{(n)} \log \hat{y}_i^{(n)} + (1-y_i^{(n)}) \log (1-\hat{y}_i^{(n)}))
\end{equation}

Now the model is able to specifically learn that an entity is not part of a particular class $i$, i.e. $w_i^{(n)}$ = 1 and $y_i^{(n)}$ = 0 for a token $t^{(n)}$ known not to belong to class $i$. This model is denoted as \textit{Sigmoid (weighted)} and is able to take advantage of all available information in the partially annotated datasets.

While this would be out-of-scope for this paper, such a model can also be used for assigning multiple labels to a single entity, in case of overlapping entity classes, or detecting nested entities\footnote{At least if the nested entity is of a different type than the entity containing it, otherwise the annotation could be ambiguous.}.

%% file: sections/4_evaluation.tex

In this Section, we provide an evaluation of all three training strategies. This includes details about the datasets we created from Wikipedia semi-automatically. 

\subsection{Datasets derived from Wikipedia}

We created two datasets, for the classes \textit{Food} and \textit{Drugs}, referring to the Wikipedia categories ``Food and Drink''\footnote{\url{https://en.wikipedia.org/wiki/Category:Food_and_drink}} and ``Drugs''\footnote{\url{https://en.wikipedia.org/wiki/Category:Drugs}}, respectively. Table \ref{tab:wiki_datasets} shows statistics about these datasets. \textit{Positive Entities}\footnote{All three models can be trained on these.} refer to the number of entities of the corresponding new class, detected through matching the set of class-related articles and hyperlinks in Wikipedia. \textit{Non-Entities}\footnote{Only \textit{Sigmoid (weighted)} can properly use these, excluded from training for \textit{Softmax (weighted)}, considered as \textit{Outside} for the \textit{Softmax} model.} denote entities that are linked to other articles in Wikipedia. \textit{Excluded Entities}\footnote{Excluded from training for the \textit{weighted} models, \textit{Outside} for the \textit{Softmax} model.} are entities that could potentially refer to an entity of the type of interest, e.g. through matching an alias of a corresponding article, but a hyperlink is missing. \textit{Entities} correspond to the number of entities for each type and \textit{Sentence} to the number of sentences in each dataset.

\begin{table}[]
\centering
\begin{tabular}{|l|l|l|l|l|l|} \hline
Entity Type       & Pos. Entities & Non-Entities & Excl. Mentions & Entities  & Sentences \\ \hline
Food              & 246,292  & 139,825      & 293,926    &  17,164          &  283,635                      \\
Drugs             &  93,439       & 16,772 &  65,350   & 27,863     &   82,498             \\ \hline  
\end{tabular}
\caption{Statistics for datasets derived from Wikipedia for the types Food and Drugs.}
\label{tab:wiki_datasets}
\end{table}

It is worth mentioning that it seems like Wikipedia knows many more drugs than food items, but still, the food-related dataset contains many more sentences. However, this is not necessarily an issue for model training, as we show in the remainder of this section.

A part of the sentences were left out for manual annotation, resulting in datasets with 280,000 and 80,000 sentences for \textit{Food} and \textit{Drugs}, respectively. 500 of each of these sets of sentences were manually annotated by an annotator familiar with the task. These datasets are referred to as \textit{Food} and \textit{Drugs gold} in the following.

\subsection{Model parameters}

We trained all models with the following settings:

\begin{itemize}
    \item Batch size: 32
    \item Optimizer: Adam \cite{kingma2014adam} with $lr=5e^{-5}$ and $\varepsilon=1e^{-8}$ without weight decay.
    \item 10 epochs, model with best F1-score on the dev dataset was selected.
    \item Train-dev-test split: 80\%-10\%-10\% with corresponding Wikipedia and CoNLL datasets merged and shuffled.\footnote{Since the CoNLL datasets could potentially contain mentions of \textit{Food} and \textit{Drugs}, the alias dictionary for each dataset was used to exclude those from training for the \textit{weighted} models.} 
\end{itemize}

The CoNLL 2003 NER dataset is split into three parts: Train, Test A (dev) and Test B (test). While Train and Test A are similar, i.e. they were extracted from news data from the same years, Test B was extracted from news articles from different years resulting in a more challenging dataset. We simply added our train, dev and test data to the appropriate CoNLL datasets.

In addition, we added a baseline approach using the alias dictionaries to annotate the manually labelled gold datasets. Each span of text appearing in these dictionaries is labelled accordingly.

\subsection{CoNLL + Wikipedia}

Table \ref{tab:conll_food} shows results (\textit{Precision}, \textit{Recall} and \textit{F1-score}) of the trained models when tested on all available datasets. The results for CoNLL test A/B can be seen as a sanity check whether the newly added data is too noisy and the output layer and loss function may or may not be able to compensate for this. In order to compare, \textit{Softmax (no food)} denotes a model, which was trained on CoNLL only, without any new Wikipedia-based datasets and entity classes.


\begin{table}[]
\centering
\begin{tabular}{|l|l|l|l|l|l|l|l|l|l|l|l|} \cline{1-7}
Dataset       & \multicolumn{3}{c|}{CoNLL test A} & \multicolumn{3}{c|}{CoNLL test B}  \\ \cline{1-7}
Model & P & R & F1 & P & R & F1 \\ \cline{1-7}
Softmax (no food) &  0.95  & 0.96 & \textbf{0.95}  & 0.90  & 0.92  & \textbf{0.91} \\
Softmax             & 0.94    &0.90 & 0.92 & 0.89 & 0.86 & 0.88  \\
Softmax (weighted)  &  0.94    &0.93 & 0.93  & 0.90 & 0.90 & 0.90  \\
Sigmoid (weighted)  & 0.93       & 0.92& 0.93 & 0.89 & 0.88 & 0.88 \\ \cline{1-7} \hline
Dataset & \multicolumn{3}{c|}{Food test A} & \multicolumn{3}{c|}{Food test B} & \multicolumn{3}{|c|}{Food gold} \\ \hline
Model & P & R & F1 & P & R & F1 & P & R & F1 \\ \hline
Softmax             & 0.78 & 0.82 & 0.80 & 0.79 &    0.82  &    0.80      &   0.73     &     0.42    & 0.53      \\
Softmax (weighted)  & 0.93 & 0.95 & \textbf{0.94} & 0.93 & 0.95     & \textbf{0.94}     &    0.52    & 0.70       & 0.59      \\
Sigmoid (weighted)  & 0.93 & 0.93 & 0.93 & 0.93 & 0.93  &    0.93      &     0.66   & 0.65         &  \textbf{0.65}  \\
Baseline & -- & -- & -- & -- & -- & -- & 0.28 & 0.51 & 0.36 \\ \hline  
\end{tabular}
\caption{Results for CoNLL 2003 mixed with the Wikipedia dataset for \textit{Food}. Best F1-score for each dataset in bold.}
\label{tab:conll_food}
\end{table}

All three approaches are able to produce reasonable results for CoNLL. Our experience was that the Wikipedia dataset for \textit{Food} did not contain a lot of entities of class \textit{Person}, \textit{Location}, \textit{Organization} or \textit{Miscellaneous}. Therefore, it is not surprising that our Wikipedia-based \textit{Food} dataset did not add too much noise harming the ability of the model to still recognize the original entity classes.

Results for \textit{Food test A/B} show the ability of the model to adapt to presumably noisy data added. The \textit{weighted} approaches clearly outperform the \textit{Softmax} approach. This shows the necessity of at least excluding mentions that are known to be ambiguous from training. The \textit{Sigmoid (weighted)} approach does not seem to add any benefit in this setting. 

The results on the \textit{Food gold} dataset are slightly more diverse. \textit{Sigmoid (weighted)} approach returned the best results with almost balanced \textit{Precision} and \textit{Recall}. Both, \textit{Softmax} and \textit{Softmax(weighted)} result in a much larger gap. The dictionary-based \textit{Baseline} approach performs poorly.

\textit{Softmax} on the one hand produces a high \textit{Precision} and low \textit{Recall}, i.e. it is capable of recognizing very few entities relatively consistently. It is possible that these entities are also often linked in Wikipedia, while others are not or at least not very often, explaining why the results on \textit{Food test A/B} do not show this phenomenon. In addition, it seems like this model pays more attention to the surface form of the mention, mainly recognizing entities it has seen before and not necessarily considering the context. 

\textit{Softmax (weighted)} on the other hand produces a high \textit{Recall} and low \textit{Precision}. This indicates that this model learned to recognize \textit{Food} entities, but since it was not trained to not recognize certain mentions that could be \textit{Food} items as well, but are for sure not (either they are linked to \textit{non-Food} Wikipedia articles or they are tagged by the CoreNLP NER), it never learned to appropriately label ambiguous spans of text. This is not an issue for \textit{Food test A/B}, presumably since (as previously mentioned), certain \textit{Food} entities are more consistently linked through hyperlinks in Wikipedia than others, while these other entities are annotated in the \textit{Food gold} dataset.

Table \ref{tab:conll_drugs} shows results for all models when trained on CoNLL 2003 and the partially annotated dataset for the class \textit{Drugs}. Overall, the results look similar to Table \ref{tab:conll_food} with similar conclusions. Test results on \textit{CoNLL test A/B} are slightly closer to the results of the model without new data added, achieving the goal of not harming the prediction quality for those classes.

For \textit{Drugs test A/B}, again, the \textit{weighted} approaches adapt a lot better to the new class than the \textit{Softmax} approach.

However, the distinction between \textit{Softmax (weighted)} and \textit{Sigmoid (weighted)} is less clear for the dataset \textit{Drugs gold}. The the latter approach has, again, a more balanced \textit{Precision} and \textit{Recall}, while the final \textit{F1-score} is the same. We assume that drug names in general are less ambiguous and therefore less non-entities are found (in fact only $\approx$10\% of all found entities from Wikipedia in the \textit{Drugs}-dataset are non-entities, compared to $\approx$20\% for the \textit{Food}-dataset). In addition, we noticed that, while manually annotating sentences for the dataset \textit{Drugs gold}, this dataset is less noisy than the \textit{Food} dataset. The \textit{Softmax} model is even able to outperform the other approaches at least on the \textit{Drugs gold} dataset, while still under-performing on \textit{Drugs test A/B}.

\begin{table}[]
\centering
\begin{tabular}{|l|l|l|l|l|l|l|l|l|l|l|l|} \cline{1-7}
Dataset       & \multicolumn{3}{c|}{CoNLL test A} & \multicolumn{3}{c|}{CoNLL test B}  \\ \cline{1-7}
Model & P & R & F1 & P & R & F1 \\ \cline{1-7}
Softmax (no drugs) &  0.95  & 0.96 & \textbf{0.95}  & 0.90  & 0.92  & \textbf{0.91} \\
Softmax             & 0.95    & 0.94 & 0.94 & 0.90 & 0.90 & 0.90  \\
Softmax (weighted)  &  0.94   &0.94 & 0.94  & 0.91 & 0.90 & \textbf{0.91}  \\
Sigmoid (weighted)  & 0.94        & 0.94 & 0.94 & 0.89 & 0.90 & 0.89 \\ \cline{1-7} \hline
Dataset & \multicolumn{3}{c|}{Drugs test A} & \multicolumn{3}{c|}{Drugs test B} & \multicolumn{3}{|c|}{Drugs gold} \\ \hline
Model & P & R & F1 & P & R & F1 & P & R & F1 \\ \hline
Softmax             & 0.89 & 0.92 & 0.90 &0.88 & 0.91 & 0.89        &     0.75   & 0.65        & \textbf{0.69}      \\
Softmax (weighted)  & 0.96 & 0.98 & \textbf{0.97} & 0.98 & 0.97     & \textbf{0.97}     &    0.58    & 0.80       & 0.67      \\
Sigmoid (weighted)  & 0.95 & 0.96 & 0.96 & 0.95 & 0.96  &    0.96      &     0.62   & 0.74         &  0.67  \\
Baseline & -- & -- & -- & -- & -- & -- & 0.28 & 0.44 & 0.34 \\ \hline  
\end{tabular}
\caption{Results for CoNLL 2003 mixed with the Wikipedia dataset for \textit{Drugs}. Best F1-score for each dataset in bold.}
\label{tab:conll_drugs}
\end{table}





%% file: sections/6_conclusion.tex

We proposed an approach to extract partially annotated datasets for Named Entity Recognition semi-automatically from Wikipedia. In addition, three model architectures, based on the commonly used BERT model, differing only in the activation function of the output layer as well as the loss function, were compared. Two of the tested models, introducing simple changes to the base model, show promising results when trained on partially annotated Wikipedia-based data and tested on similar data as well as on a small amount of manually annotated data for the classes \textit{Food} and \textit{Drugs}. Performance on the CoNLL 2003 NER dataset is not harmed significantly through adding data for the tested new entity classes. Simple dictionary-based approaches can be outperformed by a very large margin. Datasets and code are publicly available\footnote{\url{https://anonymous.4open.science/r/NER_for_partially_annotated_data-9953}}.

For future work, it would be worth investigating how to narrow the gap between test results on semi-automatically and manually annotated data. This could include a classification approach for Wikipedia articles in order to rule out articles which do not fit a certain category very well.